\documentclass[10pt,twocolumn,letterpaper]{article}

\usepackage{iccv}
\usepackage{times}
\usepackage{epsfig}
\usepackage{graphicx}
\usepackage{amsmath}
\usepackage{amssymb}
\usepackage{multirow}

\usepackage[pagebackref=true,breaklinks=true,letterpaper=true,colorlinks,bookmarks=false]{hyperref}

\iccvfinalcopy 

\ificcvfinal\fi

\begin{document}
	
	\title{Saliency-Guided Attention Network for Image-Sentence Matching}
	
\author{Zhong Ji \quad
	Haoran Wang\\  
	School of Electrical and Information Engineering\\
	Tianjin University, Tianjin, China\\
	{\tt\small  \{jizhong, haoranwang\}@tju.edu.cn}
	\and  
	Jungong Han$^*$\\	
	WMG Data Science\\ 	
	University of Warwick, Coventry, UK\\    
	{\tt\small jungong.han@warwick.ac.uk}
	\and
	Yanwei Pang\thanks{The corresponding authors are Jungong Han, Yanwei Pang.}\\
	School of Electrical and Information Engineering\\
	Tianjin University, Tianjin, China\\
	{\tt\small  pyw@tju.edu.cn}	
}
	
	%

	\maketitle

	\begin{abstract}
		This paper studies the task of matching image and sentence, where learning appropriate representations to bridge the semantic gap between image contents and language appears to be the main challenge. Unlike previous approaches that predominantly deploy symmetrical architecture to represent both modalities, we introduce a Saliency-guided Attention Network (SAN) that is characterized by building an asymmetrical link between vision and language to efficiently learn a fine-grained cross-modal correlation. The proposed SAN mainly includes three components: saliency detector, Saliency-weighted Visual Attention (SVA) module, and Saliency-guided Textual Attention (STA) module. Concretely, the saliency detector provides the visual saliency information to drive both two attention modules. Taking advantage of the saliency information, SVA is able to learn more discriminative visual features. By fusing the visual information from SVA and intra-modal information as a multi-modal guidance, STA affords us powerful textual representations that are synchronized with visual clues. Extensive experiments demonstrate SAN can improve the state-of-the-art results on the benchmark Flickr30K and MSCOCO datasets by a large margin.
	\end{abstract}
	
	\section{Introduction}
	
	Vision and language are two fundamental elements for human to perceive the real world. Recently, the prevalence of deep learning promotes them to be increasingly intertwined, which has captured great interests of researchers to explore their intrinsic correlation. In this paper, we focus on tackling the task of image-sentence matching, which facilitates various applications using cross-modal data, such as image captioning \cite{15, 54}, visual question answering (VQA) \cite{60}, and visual grounding \cite{67,68}. Concretely, it refers to searching for the most relevant images (sentences) given a sentence (image) query. Currently, the common solution \cite{9,34,24,29,33} is to seek a joint semantic space on which the data from both modalities can be well represented. Finding such a joint space is usually treated as an optimization problem where a bi-directional ranking loss encourages the corresponding representations to be as close as possible \cite{51}.

	\begin{figure}
		\begin{center}
			\includegraphics[height=3.5cm,width=6.7cm]{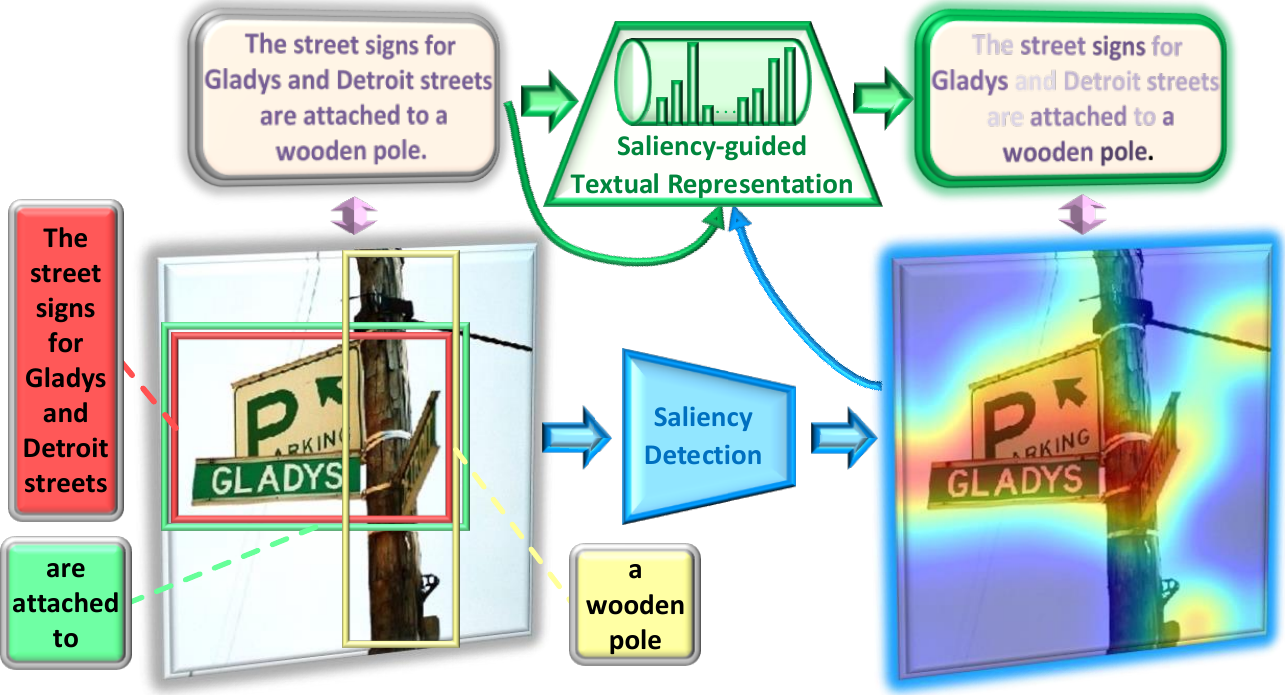}
		\end{center}
		\caption{The conceptual diagram of Saliency-guided Attention Network (SAN) for image-sentence matching. The image-sentence pair on the left denote the original data, and the colorized image regions and words on the right represent their attentive results predicted by SAN (best viewed in color).} 
		\label{fig.1}
	\end{figure}

	Although thrilling progress \cite{9,29,34,39} has been made, it is still nontrivial to represent data from different modalities in a joint semantic space precisely, due to the existence of ``heterogeneity gap''. Currently, the bulk of previous efforts \cite{9, 39} employs global Convolutional Neural Network (CNN) feature vectors as visual representations. While it could effectively represent high-level semantic information in certain tasks \cite{1, 2}, it usually makes all visual information in an image get tangled with each other. This would lead to unsatisfactory outcome because the global cross-modal similarity is obtained by aggregating the local similarities between pairwise multi-modal fragments \cite{26}. 
	
	Considering an image and its description shown in Figure \ref{fig.1}, two main objects and their relationship are: ``\emph{The street signs for Gladys and Detroit streets}'', ``\emph{a wooden pole}'' and ``\emph{are attached to}'', respectively. It indicates that, compared to the entire image, a few semantically meaningful parts may contribute more visual discrimination. In line with our observation, humans have a remarkable ability to quickly interpret a scene by selectively focusing on parts of the image rather than processing the whole scene \cite{59}. This is exactly in accordance with the purpose of visual saliency detection \cite{30, 32, 40, 42}, which aims at highlighting visually salient regions or objects in an image. Apart from the detailed structures of the objects, it also demands the understanding of the entire image \cite{13}, which is coherent with our requirement. Therefore, in this paper, we present to exploit saliency detection as guidance to benefit our visual-semantic matching model.
	
	On the other hand, grounding the representation of one modality to the finer details from the other modality plays crucial role in bridging the gap between visual and textual modality. 
	Most existing approaches \cite{18, 33, 47} adopt a two-branch symmetrical framework to represent images and sentences with the assumption that vision and language are independent and equally important. However, as the adage of ``a picture is worth a thousand words'' hints that, an image is usually more effective to convey information than a text. Inspired by this statement, our argument is that the knowledge acquired from different modalities may contribute unequally for visual-semantic matching. Specifically, the multiple sentences are potentially semantically ambiguous, owing to existence of bias and subjectivity introduced by various describers. In contrast to it, an image is not only able to provide more valuable fine-grained information but also guarantee its objectivity completely.
	Especially when considering the fact that visual saliency will further enhance visual discrimination, it is reasonable to distill knowledge from visual modality and use it to facilitate textural analysis. As illustrated in Figure \ref{fig.1}, according to the visual clues discovered by saliency detector, we take a step towards selectively attending to various words of the sentence.
	
	In this work, for addressing the issues of visual-semantic discrepancy, we propose a Saliency-Guided Attention Network (SAN) that collaboratively performs visual and textual attentions to model the fine-grained interplay between both modalities. Concretely, the SAN model is composed of three major components. A lightweight saliency detection model provides saliency with information that serves as a guidance of the subsequent two attention modules. The visual attention module selectively attends to various local visual features via resorting to the lightweight saliency detector. For the textual attention module, taking the intra-modal and inter-modal correlations into consideration, we merge the visual saliency, global visual and textual information effectively to generate multi-modal guidance, adopting soft-attention mechanism to determine the importance of word-level textual features. 
	
\begin{figure*}[!t]
		\centering
		{
			\includegraphics[height=4.95cm,width=13.8cm]{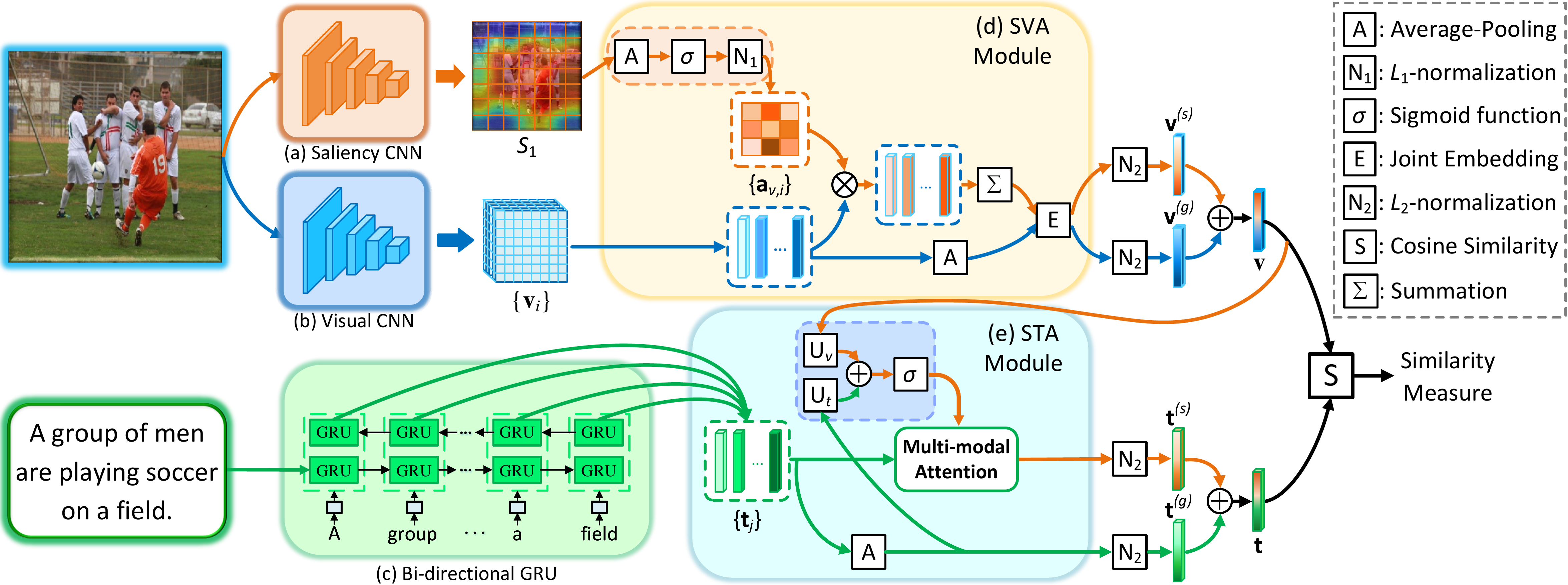}
		}
		\caption{\upshape{The proposed SAN model for image-sentence matching (best viewed in color).}}
		\label{fig.2}
	\end{figure*}

	The main contributions of our work are listed as follows:
	\begin{itemize}
		\item We propose a Saliency-Guided Attention Network (SAN), in Figure \ref{fig.2}, to simultaneously localize salient regions in an image and key words in a sentence. 
		As a departure from existing symmetric architectures that consider vision and language equally, we adopt an asymmetrical architecture emphasizing on the prior knowledge from visual modality due to the unbalanced knowledge acquired from different modalities.  
		\item A visual attention module is developed for exploiting saliency information to highlight the semantically meaningful portions of visual data and a textual attention module is presented to model the semantic interdependencies of textual data in accordance with visual information.
		\item Extensive experiments verify our SAN significantly outperforms the state-of-the-art methods on two benchmark datasets, \emph{i.e.}, MSCOCO \cite{21} and Flickr30k \cite{22}. On MSCOCO 1K test set, it improves sentence retrieval R@1 by 17.5\%. On Flickr30K, it brings about 23.7\% improvement on image retrieval R@1.
	\end{itemize}

	\section{Related Work}

	\subsection{Visual-semantic Embedding Based Image-Sentence Matching}
	
	The core idea of most existing studies \cite{9, 24, 29, 33, 34, 39, 47, 49, 50} for matching image and sentence can be boiled down to learning the joint representations for both modalities, which are roughly summarized as two main categories: 1) global alignment based methods \cite{24,29,33,39,49,50} and 2) local alignment based methods \cite{23,17,48,51,53}. Global alignment based methods usually map whole images and full sentences into a joint semantic space or learn the matching scores among pairwise multi-modal data. As a seminal work, Kiros\emph{ et al.} \cite{49} employed CNN as an image encoder and Long Short-Term Memory (LSTM) as a sentence encoder, thus constructing a joint visual-semantic embedding space with a bidirectional ranking loss. Wang \emph{et al.} \cite{33} adopted a two-layer neural network to learn structure-preserving embedding with combined cross-modal and intra-modal constraints. On the other hand, local alignment based methods usually infer the global image-sentence similarity by aligning visual objects and textual words. For instance, Karpathy \emph{et al.} \cite{23} worked on local level matching relations by performing local similarity learning among all region-words pairs. Niu \emph{et al.} \cite{48} adopted a tree-structured LSTM to learn the hierarchical relations not only between noun phrases within sentences and visual objects, but also between sentences and images. In light of the advance of object detection \cite{55}, these studies contributed to make the image-sentence matching more interpretable.
	
	To the best of our knowledge, there has been no work attempting to deploy saliency detection model to match image and sentence. Our SAN leverages it as guidance to perform attentions for both modalities, enabling us to automatically capture the latent fine-grained visual-semantic correlations.
	
	\subsection{Deep Attention Based Image-sentence Matching}
	
	The attention mechanism \cite{15} attends to certain parts of data with respect to a task-specific context, \emph{e.g.}, image sub-regions \cite{15, 54, 56} for visual attention or textual snippets for textual attention \cite{37,57}. Recently, it has been applied to conduct the image-sentence matching task. For example, Huang \emph{et al.} \cite{17} proposed a context-modulated attention to selectively focus on pairwise instances appearing in both modalities. Lee \emph{et al.} \cite{53} designed Stacked Cross Attention Network to discover the latent alignments between image regions and textual words. Unlike the above methods that explicitly aggregate local similarities to compute the global one, Nam \emph{et al.} \cite{18} developed Dual Attentional Network that performs self-attention for both modalities to capture fine-grained interplay between vision and language implicitly, which is most relevant to our work. In contrast, the major distinction lies in that our SAN constitutes an asymmetrical architecture to unidirectionally import the visual saliency information to perform textual attention learning. Doing so allows us to generate textual representations that are highly related to the corresponding visual clues.

	\section{The Proposed SAN Model}
	
	Figure \ref{fig.2} gives an overall architecture depicting our proposed SAN model. We will describe our model in detail from the following five aspects: 1) input representation for both modalities, 2) saliency-weighted visual attention with a lightweight saliency detection model, 3) saliency-guided multi-modal textual attention with the guidance of5 multi-modal information, 4) objective function for matching image and sentence, and 5) training strategy of our model.

	\subsection{Input Representation}
	
	\subsubsection{Visual Representation}
	Denote the visual features of an image $I$ by a set of convolutional features $\left\{ {\mathbf{v}_{1},...,\mathbf{v}_{M}} \right\}$, in which ${\mathbf{v}_i}\in\mathbb{R}^{d} \ ( i \in \left[ {1,M} \right] ) $ is the visual feature of the $i$-th region of images and $M$ is their total number. Specifically, given the visual features, the global visual feature $\mathbf{v}^{\left( g \right)}$ is given by:
	\begin{equation}
	\begin{aligned}
	\begin{split}
	{\mathbf{v}^{\left( g \right)}} = \mathbf{P}^{\left( g \right)}\frac{1}{M}\sum\nolimits_{i = 1}^M {{\mathbf{v}_i}},
	\end{split}
	\end{aligned}
	\end{equation}
	where matrix ${\mathbf{P}^{\left( g \right)}}$ denotes an additional fully-connected layer. It aims to embed the visual features into a $k$-dimensional joint space compatible with the textual feature.

	\subsubsection{Textual Representation}
	To build the connection between vision and language, sentence is also required to be embedded into the same $k$-dimensional semantic space. In practice, we first represent each word in a sentence with a one-hot vector, and implement word embedding on them.
	Given a sentence $T$, we split them into $L$ words $\left\{ {\mathbf{w}_{1},...,\mathbf{w}_{L}} \right\}$, and embed each word into a word embedding space with the embedding matrix $\mathbf{W}_{e}$, denoted by $\mathbf{e}_{j} = \mathbf{W}_{e}\mathbf{w}_{j}$\, ($j \in \left[ {1,L} \right]$). Then, we sequentially feed them into a bi-directional GRU at different time steps:
	\begin{equation}
	\begin{aligned}
	\begin{split}
	& {\mathbf{h}_l^f} = {GRU}^{f}(\mathbf{e}_{j}, \mathbf{h}_{j - 1}^f),\\
	& {\mathbf{h}_l^b} = {GRU}^{b}(\mathbf{e}_{j}, \mathbf{h}_{j - 1}^b),
	\end{split}
	\end{aligned}
	\end{equation}
	where $\mathbf{h}_l^f$ and $\mathbf{h}_l^b$ denote the hidden state of forward and backward GRU at time step $j$, respectively. Then, a set of textual feature vectors $\left\{ {\mathbf{t}_{1},...,\mathbf{t}_{L}} \right\}$ is the average of the forward hidden state and backward hidden state at each time step, \emph{i.e.}, $\mathbf{t}_{j} = \frac{{\mathbf{h}_l^f + \mathbf{h}_l^b}}{2}$. Specifically, given the textual feature vectors, the global textual feature vector $\mathbf{t}^{(g)}$ encoding the global information of full sentence is calculated by
	\begin{equation}
	\begin{aligned}
	\begin{split}
	{\mathbf{t}^{(g)}} = \frac{1}{L}\sum\nolimits_{j = 1}^L {\mathbf{t}_{j}}.
	\end{split}
	\end{aligned}
	\end{equation}

	\subsection{Saliency-weighted Visual Attention (SVA)}
	\label{sec:SVA}
	
	\subsubsection{The Residual Refinement Saliency Network}
	
\begin{figure}[!t]
		\centering
		\includegraphics[height=2.6cm,width=7.3cm]{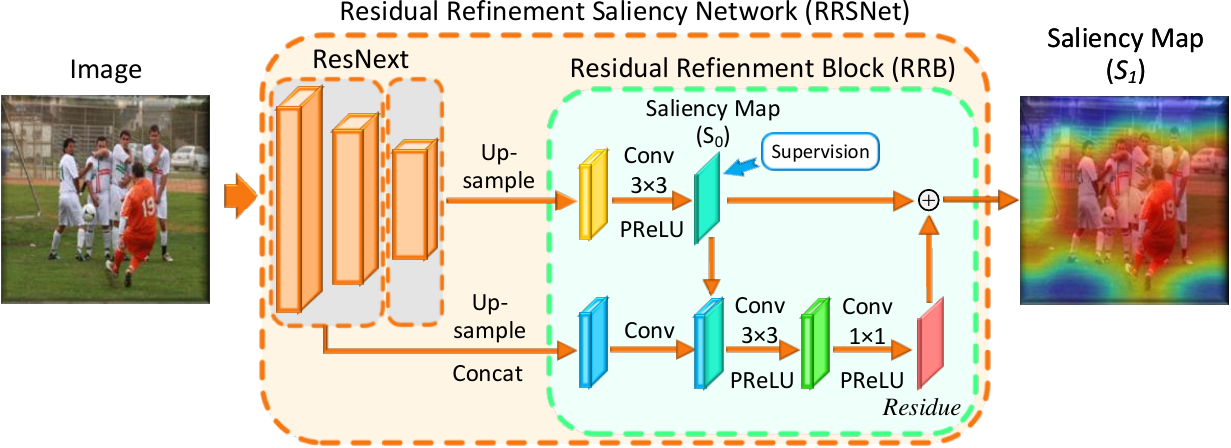}
		\caption{The architecture of RRSNet model.}
		\label{fig.3}
	\end{figure}

	Various approaches \cite{6,7,8} have been studied for visual saliency detection. But they solely focus on improving accuracy with no care about the volume of model. Hence, available visual saliency detection models are usually too big to fit into the networks for visual-semantic matching, which are usually resource restricted. Therefore, we design a lightweight saliency model, named Residual Refinement Saliency Network (RRSNet) (see Figure \ref{fig.3}), which is able to detect salient regions with limited computing resources.
	
	The implementation is as follows. First, ResNeXt-50 \cite{10} acts as the backbone network that outputs a group of feature maps with various scales. We only utilize the feature maps of the first three convolutional layers and split them into two groups: 1) low-level feature group, which contains feature maps of the first two convolutional layers; 2) high-level feature group covering the feature maps of the third convolutional layer.
	
	We first upsample the feature map of the second convolutional layer such that its size is the same as that of feature map at the first layer. Then, we concatenate them and apply the convolution operation to reduce redundant channel dimensions, producing a low-level integrated feature. We formulate this procedure as:
	\begin{equation}
	\begin{aligned}
	\begin{split}
	{F_{low}} = {g_c}(Cat({f_1},{f_2})),
	\end{split}
	\end{aligned}
	\end{equation}
	where $f_i$ denotes the feature maps being upsampled at the $i$-th convolutional layer, \emph{Cat} operation denotes concatenating feature maps at the first two convolutional layers together, $g_c(\cdot)$ is the feature fusion network that integrates low-level features via convolution operations and PReLU activation function \cite{12}. Let $F_{low}$  denote the obtained low-level integrated feature and $F_{high}$ the high-level integrated feature, we define $F_{high}$ as:
	\begin{equation}
	\begin{aligned}
	\begin{split}
	{F_{high}} = {g_c}(Cat({f_3})).
	\end{split}
	\end{aligned}
	\end{equation}

	Afterwards, we adopt the Residual Refinement Block (RRB) proposed in \cite{13}. Its principle is leveraging the low-level features and the high-level features to learn the residual between the intermediate saliency prediction and ground truth, which has been proven effective to benefit saliency predicting \cite{13}. Feeding the obtained low-level integrated feature $F_{low}$ and high-level integrated feature $F_{high}$ into the RRB, we finally acquire the refined saliency map:
	\begin{equation}
	\begin{aligned}
	\begin{split}
	& S_0 = {g_c}\left( {{F_{high}}} \right),\\
	& residue = \Phi(Cat({S_0},{F_{low}})),\\
	& S_1 = residue + {S_0},
	\end{split}
	\end{aligned}
	\end{equation}
	where $S_0$ is the initial predicted saliency map obtained by the convolution operation on $F_{high}$. Then, we feed the concatenation of $S_0$ and low-level integrated feature $F_{low}$ into function $\Phi$ to obtain the residue. Finally, the saliency map $S_1 \in \mathbb{R}^ {{H} \times {W}}$ is generated by fusing the residue and $S_0$ with element-wise addition, where $H$ and $W$ represent the height and width of input images, respectively.

	\subsubsection{Saliency-weighted Visual Attention Module}
	Distinct from most prior spatial attention schemes \cite{17, 18}, we propose to leverage the saliency information as guidance to perform visual attention, dubbed Saliency-Weighted Visual Attention (SVA). We first downsample the saliency map $S_1$ to $S_2$ with average pooling operation to align to the size of visual feature map $V \in \mathbb{R} ^{X \times Y \times d}$, which is reinterpreted as a set of $d$-dimensional visual features whose volume is $X \times Y \ (X \times Y = M)$. To preserve the spatial layout of the image, we perform average pooling over $S_1$ with a stride of $(H/X,W/Y)$. Consequently, the saliency map $S_1$ with resolution of $H \times W$ is down-sampled to match the spatial resolution of the visual feature map $V$. Then, the attention weights $\mathbf{a}_{v,i} \ (i \in \left[{1,M} \right], \sum\limits_{i = 1}^M {\mathbf{a}_{v,i}}= 1)$ can be obtained by normalizing the elements from $S_2$, achieved by applying Sigmoid function followed by ${L_1}$-normalization. Finally, with an element-wise weighted sum of visual features $\left\{ {\mathbf{v}_{i}} \right\}$ and saliency weights $\left\{ {\mathbf{a}_{v,i}} \right\}$, the salient visual feature $\mathbf{v}^{(s)}$, namely SVA vector, is calculated by:
	\begin{equation}
	\begin{aligned}
	\begin{split}
	\mathbf{v}^{(s)} = {\mathbf{P}^{(s)}}\sum\nolimits_{i = 1}^M {{\mathbf{a}_{v,i}}}  \cdot \mathbf{v}_{i},
	\end{split}
	\end{aligned}
	\end{equation}
	where $\mathbf{P}^{(s)}$ represents a fully-connected layer serving to embed visual feature into a $k$-dimensional joint space compatible with textual feature.

	\subsection{Saliency-guided Textual Attention (STA)}
	\label{sec:STA}
	
	For building the asymmetrical linking between both modalities, our scheme is resorting to attention mechanism to import the visual prior knowledge into the procedure of textual representation learning. In particular, we first merge the global visual feature $\mathbf{v}^{(g)}$ and SVA vector $\mathbf{v}^{(s)}$ into an integrated visual feature $\mathbf{v}$ with average pooling. Additionally, to make full advantage of the available semantically complementary multi-modal information, the intra-modal information $\mathbf{t}^{(g)}$ and cross-modal information $\mathbf{v}$ are further integrated to serves as the cross-modal guidance.
	
	Intuitively, a simple way to generate the multi-modal guidance is fusing the visual and textual information with element-wise addition. However, it may lead to valid information in one modality be concealed by the other one. To alleviate this issue, we design a gated fusion unit to combine them effectively. Specifically, given the integrated visual feature $\mathbf{v}$ and the global textual feature $\mathbf{t}^{(g)}$, we feed them into the gated fusion unit, which is formulated as:
	\begin{equation}
	\begin{aligned}
	\begin{split}
	& \hat {\mathbf{v}} = \mathbf{U}_{v} ( {\mathbf{v}} ), \hat {\mathbf{t}} = \mathbf{U}_{t} ( {\mathbf{t}}^{(g)} ), \hfill \\
	& {\mathbf{m}}_{f} = \sigma ( \hat {\mathbf{v}} + \hat {\mathbf{t}} ), \hfill
	\end{split}
	\end{aligned}
	\end{equation}
	where $\mathbf{U}_{v}$ and $\mathbf{U}_{t}$ denote parameters of two fully-connected layers, respectively. The Sigmoid function $\sigma$ is used to rescale each element in the fused representation to $[0,1]$. $\mathbf{m}_{f}$ represents the refined multi-modal context vector output by the gated fusion unit.
	
	Then, we leverage the soft attention mechanism to perform textual attention. Specifically, given the refined multi-modal context $\mathbf{m}_{f}$ and the textual feature ${\mathbf{t}_{j}}\ (j \in \left[ {1,L} \right])$, the saliency-guided textual feature vector ${\mathbf{t}^{(s)}}$, namely STA vector, is calculated by:
	
	\begin{equation}
	\begin{aligned}
	\begin{split}
	& {\mathbf{h}_{t,j}} = \tanh ({\mathbf{W}}_t^{(0)}{\mathbf{m}_f}) \odot \tanh ({\mathbf{W}}_t^{(1)}{\mathbf{t}_j}), \\
	& {\mathbf{a}_{t,j}} = {\rm softmax } ({\mathbf{W}}_t^{(2)}{\mathbf{h}_{t,j}}), \\
	& {\mathbf{t}^{(s)}} = \sum\nolimits_{j = 1}^L {{\mathbf{a}_{t,j}} \cdot {\mathbf{t}_j}},
	\end{split}
	\end{aligned}
	\end{equation}
	where ${\mathbf{W}}_t^{(0)}$, ${\mathbf{W}}_t^{(1)}$ and ${\mathbf{W}}_t^{(2)}$ are parameters of three fully-connected layers, respectively. ${\mathbf{h}_{t,j}}$ denotes the hidden state of textual attention and ${\mathbf{a}_{t,j}} \ (j \in \left[ {1,L} \right])$ is the textual attention weight. Similar to visual modality, we obtain the integrated textual feature $\mathbf{t}$ via merging the global textual feature ${\mathbf{t}^{(g)}}$ and STA vector ${\mathbf{t}^{(s)}}$ with average pooling.


	\subsection{Objective Function}
	We follow \cite{23,26,33} to employ the bidirectional triplet loss, which is defined as:
	\begin{equation}
	\begin{aligned}
	\begin{array}{l}
	\mathcal{L}_{rank}\left( {I,T} \right) = \sum\limits_{(I,T)} \{  \max [0,\gamma  - s(I,T) + s({I^ - },T)] \\
	\begin{array}{*{10}{c}}
	{\begin{array}{*{10}{c}}
		{\begin{array}{*{10}{c}}
			{}&{}
			\end{array}}&{}
		\end{array}}&{}
	\end{array}
	+ \max [0,\gamma - s(I,T) + s(I,{T^ - })]\},
	\end{array}
	\end{aligned}
	\end{equation}
	where $\gamma$ denotes the margin parameter and $s\left( { \cdot , \cdot } \right)$ denotes the Cosine function. Given a matched image-sentence pair $(V,T)$, its corresponding negative samples are denoted as $V^-$ and $T^-$, respectively.

	\subsection{Two-Stage Training Strategy}
	
	The training procedure of our proposed SAN model involves two stages. We first train the proposed RRSNet, while freezing the parameters of the remaining part of the SAN. The MSRA10K dataset \cite{30} serves as supervision for training RRSNet, which is widely utilized in saliency detection \cite{31, 32}. Similar to the extensive application of pre-trained CNN for visual representation, we first train the RRSNet alone aming at importing its available prior knowledge of saliency to benefit the training procedure of next stage. After stage-1 converges, we start stage-2 for fine-tuning the parameters of the whole SAN model.

	\section{Experimental Results and Analyses}
	
	To verify the effectiveness of the proposed SAN model, we carry out extensive experiments in terms of image retrieval and sentence retrieval on two publicly available benchmark datasets: MSCOCO \cite{21} and Flickr30K \cite{22}.
	
	\subsection{Datasets and Evaluation Metrics}
	
	\textbf{Datasets}. The two datasets and their corresponding experimental protocols are introduced as follows: \textbf{1) MSCOCO} \cite{21} consists of 123,287 images, and each image contains roughly five textual descriptions. It is split into 113,287 training images, 5,000 validation images and 5,000 testing images \cite{23}. The experimental results are reported by averaging over 5-fold cross-validation.
	\textbf{2) Flickr30k} \cite{52} contains 31,783 images collected from the Flickr website, in which each image is annotated with five caption sentences. Following \cite{23}, we split the dataset into 29,783 training images, 1000 validation images and 1000 testing images.
	
	\textbf{Evaluation Metrics.} We use two evaluation metrics, \emph{i.e.}, R@K (K=1,5,10) and ``mR''. R@K denotes the percentage of ground-truth matchings appearing in the top K-ranked results. Besides, we also follow \cite{26} to adopt average of all six recall rates of R@K to obtain ``mR'', which is more reasonable to evaluate the overall performance for cross-modal retrieval.

	\begin{table*}[!t]
		
		\normalsize
		
		\begin{center}
			\caption{Comparisons of experimental results on MSCOCO 1K test set and Flickr30k test set. The visual feature extractors of all methods are provided for reference.}
			\label{tab.1}
			
			\resizebox{0.9\textwidth}{2.9cm}{
				\begin{tabular}{l|ccc|ccc|c|ccc|ccc|c}
					\hline
					\hline
					\multicolumn{1}{c|}{\multirow{3}{*}{Approach}}  & \multicolumn{7}{c|}{ MSCOCO  dataset}                                                                       & \multicolumn{7}{c}{Flickr30k dataset}                                                                                                                                                   \\
					\hline
					& \multicolumn{3}{c|}{Sentence Retrieval}                                       & \multicolumn{3}{c|}{Image Retrieval}                                          & \multicolumn{1}{c|}{\multirow{2}{*}{mR}} & \multicolumn{3}{c|}{Sentence Retrieval}                                       & \multicolumn{3}{c}{Image Retrieval}
					& \multicolumn{1}{|c}{\multirow{2}{*}{mR}}\\
					
					\multicolumn{1}{c}{}                           & \multicolumn{1}{|c}{R@1} & \multicolumn{1}{c}{R@5} & \multicolumn{1}{c|}{R@10} & \multicolumn{1}{c}{R@1} & \multicolumn{1}{c}{R@5} & \multicolumn{1}{c|}{R@10} & \multicolumn{1}{c|}{}   & \multicolumn{1}{c}{R@1} & \multicolumn{1}{c}{R@5} & \multicolumn{1}{c|}{R@10} & \multicolumn{1}{c}{R@1} & \multicolumn{1}{c}{R@5} & \multicolumn{1}{c|}{R@10} & \multicolumn{1}{c}{}   \\
					\hline
					\hline
					DVSA \cite{23} (R-CNN)        & 38.4                    & 69.9                    & 80.5                     & 27.4                    & 60.2                    & 74.8                     & 39.2                   & 22.2                    & 48.2                    & 61.4                     & 15.2                    & 37.7                    & 50.5                     & 58.5                   \\
					
					\hline
					
					m-RNN \cite{34} (VGG-16)      &	41.0 &	73.0 &	83.5 &	29.0 &	42.2 &	77.0  &	57.6 &	35.4 & 63.8 & 73.7 & 22.8 & 50.7 & 63.1 & 51.6  \\
					
					GMM-FV \cite{29} (VGG-16)    & 35.0                    & 62.0                    & 73.8                     & 25.0                    & 52.7                    & 66.0                     & 52.4                   & 39.4                    & 67.9                    & 80.9                     & 25.1                    & 59.8                    & 76.6                     & 58.3                   \\
					
					m-CNN \cite{39} (VGG-19)        & 33.6                    & 64.1                    & 74.9                     & 26.2                    & 56.3                    & 69.6                     & 54.1                   & 42.8                    & 73.1                    & 84.1                     & 32.6                    & 68.6                    & 82.8                     & 64                     \\
					
					DSPE \cite{33} (VGG-19) & 40.3                    & 68.9                    & 79.9                     & 29.7                    & 60.1                    & 72.1                     & 58.5                   & 50.1                    & 79.7                    & 89.2                     & 39.6                    & 75.2                    & 86.9                     & 70.1                   \\
					
					2WayNet \cite{65} (VGG-19)   & 49.8                    & 67.5                    & -                        & 36.0                    & 55.6                    & -                        & -                      & 55.8                    & 75.2                    & -                        & 39.7                    & 63.3                    & -                        & -                      \\
					
					\hline
					
					CMPM \cite{58} (ResNet-152)  & 56.1 &	86.3 &	92.9  &	44.6 &	78.8 &	89  & 74.6   & 49.6 &	76.8 &	86.1  &	37.3 &	65.7 &	75.5  &	65.2  \\
					
					VSE++ \cite{9} (ResNet-152)     & 64.7                    & -                       & 95.9                     & 52.0                    & -                       & 92.0                     & -         & 52.9                    & -                       & 87.2                     & 39.6                    & -                       & 79.5                     & -              \\
					
					DPC \cite{47} (ResNet-50)    & 	65.6 &	89.8 &	95.5  &	47.1 &	79.9 &	90.0  &	78.0  &  55.6 &	81.9 &	89.5 &	39.1 &	69.2 &	80.9  &	69.4 \\
					
					PVSE \cite{66} (ResNet-152)  & 69.2                   & 91.6                      & 96.6                     & 55.2                       & 86.5                   & 93.7	& -  & -  & -  & -  & -  & -  & -                \\
					
					SCO \cite{26} (ResNet-152)  & 69.9 &	92.9 &	97.5  &	56.7 &	87.5 &	94.8 &	83.2 & 55.5 &	82.0 &	89.3  &	41.1 &	70.5 &	80.1  &	69.7  	\\
					
					\hline
					
					SCAN \cite{53} (Faster R-CNN)   & 72.7 &	94.8 &	98.4 &	58.8 &	88.4 &	94.8  &	83.6   &	67.4 &	90.3 &	95.8 &	48.6 &	77.7 &	85.2 &	77.5	\\
					
					\hline \hline
					SAN (VGG-19)
					&	74.9 &	94.9 &	98.2  &	60.8 &	90.3 &	95.7 &	85.8
					&	67.0 &	88.0 &	94.6 &	51.4 &	77.2 &	85.2 &	77.2 \\
					SAN (ResNet-152)
					& \textbf{85.4} &	\textbf{97.5} &	\textbf{99} &	\textbf{69.1}
					& \textbf{93.4} &	\textbf{97.2}  & \textbf{90.3}  &	\textbf{75.5} &	\textbf{92.6} &	\textbf{96.2}  &	 \textbf{60.1} &	\textbf{84.7} &	\textbf{90.6}  & \textbf{83.3}  	\\
					\hline
					\hline
					
				\end{tabular}
			}
		\end{center}
		
	\end{table*}

	\subsection{Implementation Details}
	
	All our experiments are implemented in pytorch toolkit with a single NVIDIA GEFORCE GTX TITAN Xp GPU. As for image preprocessing, we first resize the input image to 256$\times$256, and then follow \cite{29} to use the average of the feature vectors for 10 crops of size 224$\times$224. In this paper, we elect the last pooling layer of ResNet-152 (res5c) as the visual feature. Additionally, we also provide the results by using the last pooling layer of VGG-19 (pool5) for comparison. The size of visual feature map output by image encoder is 7$\times$7$\times$512 for VGG-19 and 7$\times$7$\times$2048 for ResNet-152, respectively. The dimensionality of word embedding space is set to 300, and that of the bi-directional GRU units and the joint space $k$ is set to 1024. The margin parameter $\gamma$ is empirically set to 0.2.
	As mentioned above, the training procedure includes two stages. At the first stage, the saliency model RRSNet is trained for 8000 iterations by the stochastic gradient descent (SGD) with mini-batch size of 96, setting the leaning rate to 0.001. At the second stage, we train our SAN model by Adam optimizer \cite{27} with mini-batch size of 128 and fixed learning rate of 0.00005.

	\begin{table}[!t]
		
		\huge
		
		\begin{center}
			\setlength{\abovecaptionskip}{0pt}%
			\setlength{\belowcaptionskip}{20pt}%
			
			\caption{Comparisons of experimental results on MSCOCO 5K test set. The visual feature extractors of all methods are provided for reference.}
			\label{tab.2}
			\resizebox{0.47\textwidth}{1.86cm}{
				\begin{tabular}{l|ccc|ccc|c}
					\hline
					\hline
					\multicolumn{1}{c|}{\multirow{2}{*}{Approach}}
					&	\multicolumn{3}{|c}{Sentence Retrieval} & \multicolumn{3}{|c}{Image Retrieval} & \multicolumn{1}{|c}{\multirow{2}{*}{mR}} 	\\
					&   R@1 & R@5 & R@10  & R@1 & R@5 & R@10  & \multicolumn{1}{c}{} \\
					\hline
					\hline
					\multicolumn{7}{c}{MSCOCO dataset (5K test set)} \\
					\hline
					
					DVSA \cite{23} (R-CNN)     &	16.5 &	39.2 &	52.0  &	10.7 &	29.6 &	42.2  & 31.7 \\
					
					\hline
					
					VSE++ \cite{9} (ResNet-152)		&	41.3 &	- &	81.2 &	30.3 &	- &	72.4 &	- \\
					
					DPC \cite{47} (ResNet-152)  & 	41.2 &	70.5 &	81.1  &	25.3 &	53.4 &	66.4   &	56.3\\
					
					GXN \cite{50} (ResNet-152)  & 	42.0 &	-    &	 84.7 &	31.7 &	-    &  74.6  &	- \\
					
					SCO \cite{26} (ResNet-152)  &	42.8 &	72.3 &	83.0  &	33.1 &	62.9 &	75.5  & 61.6 \\
					
					PVSE \cite{66}  (ResNet-152)	   &   45.2   &  74.3 &  84.5 &  32.4 &  63.0	&  75.0  & 62.4 \\
					
					\hline
					
					SCAN \cite{53} (Faster R-CNN)      &	50.4 &	82.2 &	90  & 38.6 &	69.3 &	80.4  &	68.5 \\
					\hline\hline
					
					SAN (ResNet-152) & \textbf{65.4} & \textbf{89.4} & \textbf{94.8}  &\textbf{46.2} & \textbf{77.4} & \textbf{86.6} & \textbf{76.6} \\	
					\hline
					\hline
				\end{tabular}
			}
		\end{center}
		
	\end{table}

	\begin{figure*}[!t]
		\centering
		\includegraphics[width=0.85\linewidth,height=6.8cm]{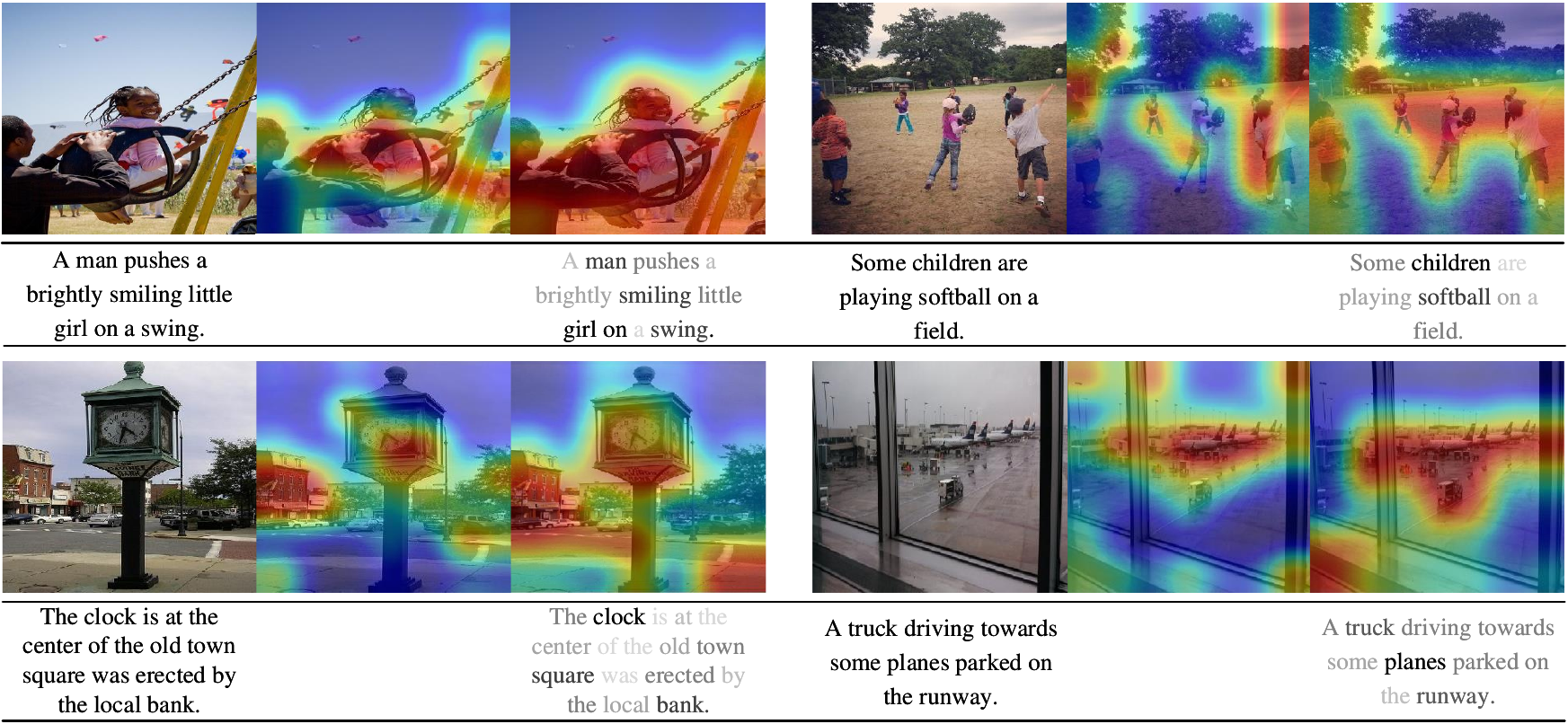}
		\caption{Attention visualization on MSCOCO dataset. The original image, the saliency heatmaps produced by Stage-1 training and Stage-2 training are shown from left to right, respectively. Their corresponding description are shown below them (Best viewed in color).}
		\label{fig.4}
	\end{figure*}

	\subsection{Comparisons with the State-of-the-art Approaches}
	We compare our proposed SAN model with several state-of-the-art approaches on MSCOCO and Flickr30k datasets for bidirectional image and sentence retrieval, respectively.

	\subsubsection{Results on MSCOCO Dataset}
	The experimental results on the MSCOCO 1K and 5K test set are shown in Table \ref{tab.1} and Table \ref{tab.2}, respectively. From Table \ref{tab.1}, we can observe that our SAN model significantly outperforms all competitors in all nine evaluation metrics, which clearly demonstrates the superiority of our approach. Our best result on 1K test set is achieved by employing the ResNet-152 as the visual feature. Take R@1 for example, there are 12.7\% and 10.3\% improvements against the second best SCAN approach \cite{53} on sentence retrieval and image retrieval, respectively. Moreover, it is apparent that even with the VGG-19 as image encoder, our model also exceeds the best competitor by 2.2\% on mR. Besides, as illustrated in Table \ref{tab.2}, compared to other baselines, we achieve considerable boost of 15\% on R@1 for sentence retrieval and 7.6\% on R@1 for image retrieval on the 5K test set.

	\subsubsection{Results on Flickr30k Dataset}
	The results on the Flickr30K dataset is listed in Table \ref{tab.1}. In the case of adopting VGG-19 as image encoder, our proposed SAN achieves competitive performance, improving 11.5\% comparing to SCO on R@1 for sentence retrieval. Furthermore, our best result based on ResNet-152 achieve new state-of-the-art performance and yield a result of 75.5\% and 60.1\% on R@1 for sentence retrieval and image retrieval, respectively. Comparing with the best competitor, we achieve absolute boost of 8.1\% on R@1 for sentence retrieval and 11.5\% on R@1 for image retrieval.

	\subsection{Ablation Studies}
	
	\subsubsection{Ablation Models for Comparisons}
	\label{sec.ablation}
	In this section, we perform several ablation studies to systematically explore the impacts of both attention modules of SAN. Thus we develop various ablation models
	and display their configurations in Table \ref{tab.3}. To focus on studying the impact brought by the SVA module and STA module, we configure the SAN model with various representation components for both modalities. For visual representation,  the variable representation methods are illustrated as follows: 1) ``GV'' and ``SV'' denote employing the global visual feature ${\mathbf{v}^{(g)}}$ and SVA feature vector ${\mathbf{v}^{(s)}}$ as visual representations, respectively. 2) ``FV'' refers to utilize the fused visual feature ${\mathbf{v}}$ acquired by fusing ``GV'' and ``SV'' feature together, as mentioned in section \ref{sec:SVA}. For textual modality, the variable representation methods are explained as follows:  3) ``GT'' and ``ST'' indicate deploying the global textual feature ${\mathbf{t}^{(g)}}$ and the STA vector ${\mathbf{t}^{(s)}}$ as textual representations, respectively. 4) ``IT'' means adopting the attentive textual feature generated by leveraging the self-attention mechanism proposed in \cite{18} as textual representation. 5) ``FT (G-I)'' is the integrated textual feature obtained by merging ``GT'' and ``IT'' feature together with average pooling. 6) ``FT (G-S)'' denotes the fused textual feature ${\mathbf{t}}$ that is described in section \ref{sec:STA}. Note that, in the following ablation  experiments, we validate the performance on the 1K test set of MSCOCO and adopt ResNet-152 as our default image encoder.

	\begin{table}
		
		\Huge
		
		\begin{center}
			\caption{The ablation models with different experimential settings.}
			\label{tab.3}
			\resizebox{0.46\textwidth}{1.88cm}{
				\begin{tabular}{l|cc|ccc}
					\hline \hline
					\multicolumn{1}{c}{\multirow{2}{*}{Ablation Model}} &
					\multicolumn{2}{|c}{Visual Representation} & \multicolumn{3}{|c}{Textual Representation}  \\
					& $\quad$ GV  & $\quad$SV 	& $\quad$ GT 	& \ $\quad$ IT 	& $\quad$ ST  \\
					\hline \hline
					SAN (GV + GT) &	$\quad$ $\checkmark$ & $\quad$ & \  \ $\quad$$\checkmark$ &	 & 	\\
					\hline
					SAN (SV \,+ GT) &	 &	\ $\quad$$\checkmark$ &	\  \ $\quad$$\checkmark$ &	&  	\\
					SAN (FV \,+ IT) &	$\quad$	$\checkmark$  &	\ $\quad$$\checkmark$ &	\  &	\ \ $\quad$$\checkmark$ &	\\
					\hline \hline
					SAN (FV \,+ GT) &	$\quad$	$\checkmark$  &	\ $\quad$$\checkmark$ &	\ \ $\quad$$\checkmark$  & \	 &	\\
					\hline \hline
					SAN (FV \,+ FT(G-I)) &	\ $\quad$$\checkmark$  &	\ $\quad$$\checkmark$ & \  \ $\quad$$\checkmark$ 	&	\ \ $\quad$$\checkmark$ &	\\
					SAN (GV + FT(G-S))  & \ $\quad$$\checkmark$  &	\ $\quad$ & \  \ $\quad$$\checkmark$ 	&	 & \ \ $\quad$$\checkmark$	\\
					SAN (SV \,+ FT(G-S))  &   &	\ $\quad$$\checkmark$ & \  \ $\quad$$\checkmark$ 	&	 & \ \ $\quad$$\checkmark$	\\
					SAN (FV \,+ FT(G-S))  & \ $\quad$$\checkmark$  & \ $\quad$$\checkmark$ & \  \ $\quad$$\checkmark$ 	&	 & \ \ $\quad$$\checkmark$	\\
					\hline \hline
				\end{tabular}
			}
		\end{center}
	\end{table}

	\begin{table}
		\Huge
		\begin{center}
			\caption{Impact of single SVA module on MSCOCO 1K test set.}
			\label{tab.4}
			\resizebox{6.25cm}{1.15cm}{
				\begin{tabular}{l|cc|cc}
					\hline \hline
					\multicolumn{1}{c|}{\multirow{2}{*}{Approach}} &
					\multicolumn{2}{c}{Sentence Retrieval} & \multicolumn{2}{|c}{Image Retrieval}  \\
					& R@1 & R@10 & R@1 & R@10 \\
					\hline \hline
					SAN (GV + GT) &	63.4 &	92.8 &	50.5 &	88.8 	\\
					SAN (SV \,+ GT) &	65.4 &	95.1 &	52.9 &	91.1 	\\
					SAN (FV \,+ GT) &	66.2 &	95.7 &	53.7 &	92.0 	\\
					SAN (FV \,+ IT) &	\textbf{67.1}  & \textbf{96.6} &	\textbf{56.6} &	\textbf{93.5}  \\
					\hline \hline
				\end{tabular}
			}
		\end{center}
	\end{table}

	\subsubsection{Evaluating the Impact of SVA Module}
	
	To systematically explore the contribution of the SVA module, we specially remove the STA module from the entire SAN model. Specifically, we select the following four ablation models illustrated in section \ref{sec.ablation} to validate the effectiveness of the SVA component: SAN (GV + GT), SAN (SV \,+ GT), SAN (FV \,+ GT) and SAN (FV \,+ IT).
	
	Taking SAN (GV + GT) model as baseline, we can obtain the following conclusions from Table \ref{tab.4}: 1) Replacing the global visual feature (``GV'') with SVA feature (``SV'') will provide additional 2.0\% improvement for sentence retrieval and 2.4\% improvement for image retrieval on R@1, respectively. 2) Fusing the global visual feature (``GV'') with SVA feature (``SV'') together yields better results, indicating the SVA module and the image encoder are mutually beneficial for enhancing the discrimination of visual modality. 3) When deploying a better textual feature (``IT'') that benefits from the self-attention mechanism \cite{37}, the SVA module can still collaborate with it, resulting in better performance. These results verify that our proposed SVA module has capability to boost retrieval performance independently, indicating the saliency information are actually conducive to understanding an image semantically.

	\subsubsection{Evaluating the Impact of STA Module}
	
	To further delve into the effect of STA module, we perform two groups of ablation experiments, and shown the results in Table \ref{tab.5} and Table \ref{tab.6}, respectively. First, we focus on exploring the impact of various textual representations. From Table \ref{tab.5}, we see that the performance gain brought by combining the global textual feature (``GT'') with attentive textual feature (``IT'') is really slight. By contrast, the significant performance gain by only equipping the baseline (SAN (GV + GT)) with the STA module can be observed, bringing about 19.2\% improvement on R@1 for sentence retrieval and 15.4\% improvement on R@1 for image retrieval. The experimental results validate demonstrate the superiority of our proposed STA module.
	
	Moreover, we investigate the influence of feeding various visual information into the STA module. From Table \ref{tab.6}, it is worth noting that even if the SVA feature (``SV'') is excluded, implying no saliency information contained in the visual modality, our SAN (GV + FT(G-S)) is still comparable to the current state-of-the-art \cite{53}. Besides, we see that the performance gain acquired by replacing ``GV'' with ``SV'' feature is very compelling. It strengthens our belief that the saliency information plays crucial role in bridging the gap between two separate modalities and transferring more effective information to textual modality.

	\begin{table}
		\Huge
		\begin{center}
			\caption{Impact of different textual representations on MSCOCO 1K test set.}
			\label{tab.5}
			\resizebox{6.55cm}{0.90cm}{
				\begin{tabular}{l|cc|cc}
					\hline \hline
					\multicolumn{1}{c|}{\multirow{2}{*}{Approach}} &
					\multicolumn{2}{c|}{Sentence Retrieval} & \multicolumn{2}{c}{Image Retrieval}  \\
					& R@1  & R@10 & R@1 & R@10 \\
					\hline \hline
					SAN (FV \,+ GT) 	 & 66.2  &	95.7 &	53.7   &	92.0  \\
					SAN (FV \,+ FT(G-I)) &	66.8  &	96.5 &	55.8 &	93.6  \\
					SAN (FV \,+ FT(G-S)) &	\textbf{85.4}  & \textbf{99} & \textbf{69.1}  & \textbf{97.2} \\
					\hline \hline
				\end{tabular}
			}
		\end{center}

		\begin{center}
			\caption{Impact of variable guidance information of STA module on MSCOCO 1K test set.}
			\label{tab.6}
			\resizebox{6.55cm}{0.90cm}{
				\begin{tabular}{l|cc|cc}
					\hline \hline
					\multicolumn{1}{c|}{\multirow{2}{*}{Approach}} &
					\multicolumn{2}{c|}{Sentence Retrieval} & \multicolumn{2}{c}{Image Retrieval}  \\
					& R@1  & R@10 & R@1 & R@10 \\
					\hline \hline
					SAN (GV + FT(G-S))  &	74.5  &	97.8 &	57.8 &	94.6  \\
					SAN (SV \,+ FT(G-S))  &	82.1  &	97.8 &	67.3 &	96.4  \\
					SAN (FV \,+ FT(G-S))  &	\textbf{85.4}  & \textbf{99} & \textbf{69.1}  & \textbf{97.2} \\
					\hline \hline
				\end{tabular}
			}
		\end{center}
	\end{table}

	\subsection{Qualitative Results and Analysis}
	
	The visualization of attention outputs on MSCOCO are depicted in Figure \ref{fig.4}. The two attention heatmaps correspond to the predicted saliency maps generated by SVA module after stage-1 and stage-2 training, respectively. As illustrated in Figure \ref{fig.4}, the salient regions detected after Stage-1 training appear coarse-grained and incompletely consistent with our common sense. In comparison, the attention heatmaps from Stage-2 training contain more interpretable fine-grained visual clues contributing to understand the image from an overall perspective, which seems more in accordance with human intuition. More concretely, taking the case of the fourth selected image, the SVA module (Stage-2) can not only focus on most salient objects ``plan'' just as it achieves in stage-1, but also allocate enough attention on the ``trunk'' and ``runway''. Similarly, the textual attention weights output by STA module appear reasonable as well. These observations demonstrate our SAN succeed to learn interpretable alignments between image regions and words, meanwhile allocates reasonable attention weights to the visual regions and textual words according to their respective semantic importance.
	
	\begin{figure}[!t]
		\centering
		\includegraphics[height=4.4cm, width=0.9\linewidth]{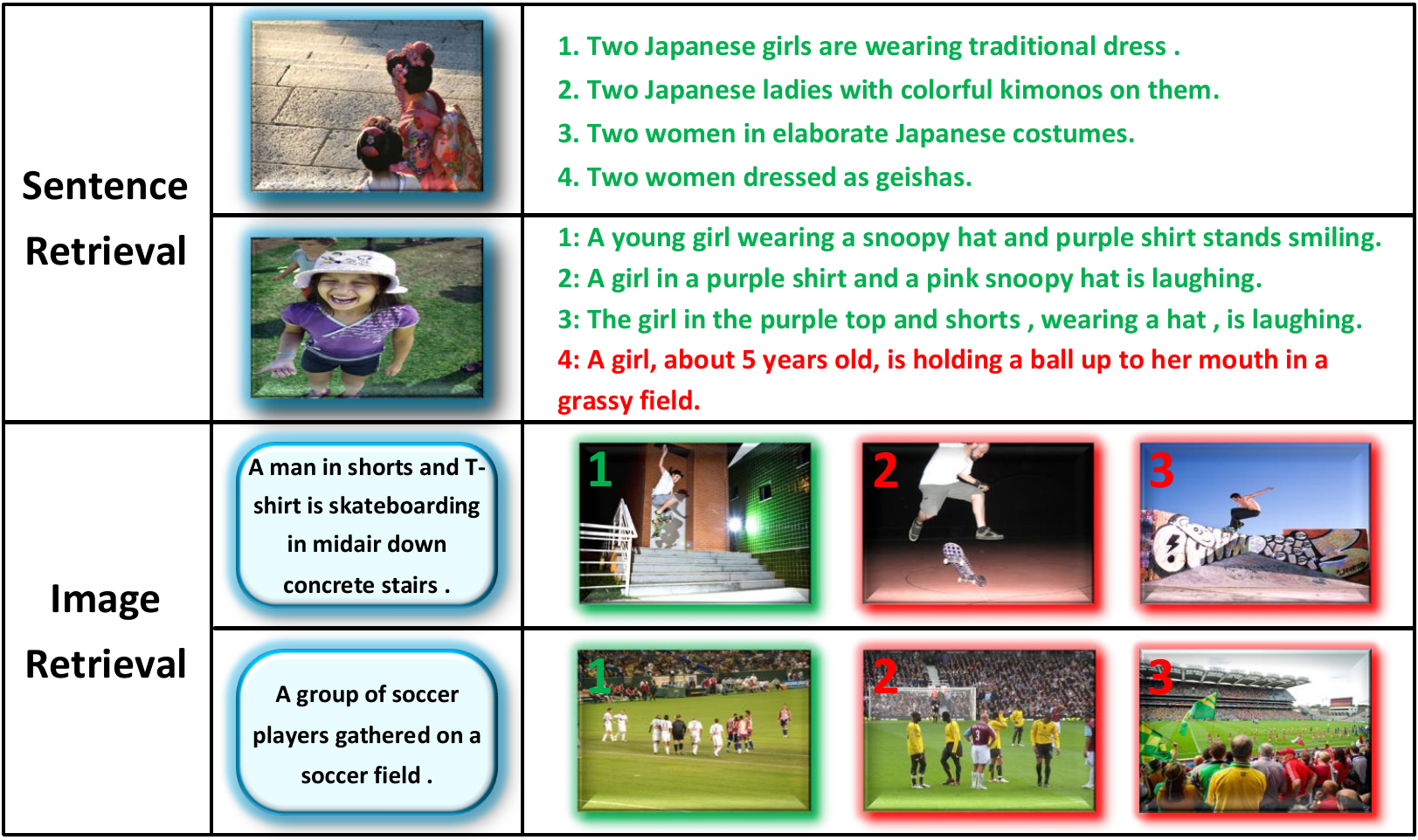}
		\caption{Qualitative results of bidirectional retrieval on Flickr30K dataset. For each image query, the top 4 corresponding ranked sentences are presented. For each sentence query, we present the top 3 ranked images, ranking from left to right. We mark the matched results in green and mismatched results in red (Best viewed in color).}  
		\label{fig.5}
		~\\
	\end{figure}
	
	To further qualitatively verify the effectiveness of SAN, we select several representative images and sentences to show their corresponding retrieval results on Flickr30k in Figure \ref{fig.5}, respectively. We observe SAN returns the reasonable retrieval results.

	\section{Conclusion}
	In this work, we proposed a Saliency-Guided Attention Network (SAN) for matching image and sentence, which is characterized by employing two proposed attention modules to associate both modalities with asymmetric fashion. Specifically, we introduce a spatial attention module and a textual attention module to capture the fine-grained cross-modal correlation between image and sentence. The ablation experiments exhibit the two attention modules are not only capable of boosting retrieval performance individually, but also complementary and mutually beneficial to each other. Experimental results on Flickr30K and MSCOCO datasets demonstrate our SAN considerably exceed the state-of-the-art by a large margin.

	\section{Acknowledgement}
	This work was supported by the National Natural Science Foundation of China under Grants 61771329 and 61632018.

	{\small
		\bibliographystyle{ieee_fullname}
		\bibliography{Reference-SAN_fullname}
	}

\end{document}